\documentclass[journal]{IEEEtran}
\usepackage{amsmath,amsfonts}
\usepackage{algorithmic}
\usepackage{array}
\usepackage[caption=false,font=normalsize,labelfont=sf,textfont=sf]{subfig}
\usepackage{textcomp}
\usepackage{stfloats}
\usepackage{url}
\usepackage{cite}
\usepackage{booktabs}
\usepackage{multirow}
\usepackage{makecell}
\usepackage{verbatim}
\usepackage{graphicx}
\usepackage[hidelinks]{hyperref} % 先保证不出框
\hypersetup{
  colorlinks=true,
  citecolor=blue,
  linkcolor=blue,
  urlcolor=blue
}
\def\BibTeX{{\rm B\kern-.05em{\sc i\kern-.025em b}\kern-.08em
    T\kern-.1667em\lower.7ex\hbox{E}\kern-.125emX}}

\begin{document}
\title{DUGAE: Unified Geometry and Attribute Enhancement via Spatiotemporal Correlations for G-PCC Compressed Dynamic Point Clouds}
\author{Pan Zhao, Hui Yuan,~\IEEEmembership{Senior Member,~IEEE,} Chang Sun, Chongzhen Tian, Raouf Hamzaoui,~\IEEEmembership{Senior Member,~IEEE,} and Sam Kwong,~\IEEEmembership{Fellow,~IEEE}

\thanks{This work was supported in part by the National Natural Science Foundation
of China under Grants 62222110, 62571303, 62172259, and 62072150, the Foreign Experts Recruitment Plan of Chinese Ministry of Human Resources and Social Security under Grant H20251083, the Taishan Scholar Project of Shandong Province (tsqn202103001), the Shandong Provincial Natural Science Foundation under Grant ZR2022ZD38. (Corresponding author: Hui Yuan)}

\thanks{Pan Zhao, Hui Yuan, Chang Sun, and Chongzhen Tian are with the School of Control Science and
Engineering, Shandong University, Jinan 250061, China, and also with
the Key Laboratory of Machine Intelligence and System Control, Ministry
of Education, Ji’nan, 250061, China (e-mail: panz@mail.sdu.edu.cn;
huiyuan@sdu.edu.cn; vimsunchang@mail.sdu.edu.cn; 202420789@mail.sdu.edu.cn)

Raouf Hamzaoui is with the School of Engineering, Infrastructure and Sustainability De Montfort University, LE1 9BH Leicester, UK. (e-mail:
rhamzaoui@dmu.ac.uk)

Sam Kwong is with the Department of Computing and Decision Science,
Lingnan University, Hong Kong. (e-mail: samkwong@ln.edu.hk),

}}

\markboth{Journal of \LaTeX\ Class Files,~Vol.~18, No.~9, January~2026}%
{DUGAE: Unified Geometry and Attribute Enhancement via Inter-Frame Spatiotemporal Correlations for G-PCC Compressed Dynamic Point Clouds}

\maketitle

\begin{abstract}
Existing post-decoding quality enhancement methods for point clouds are designed for static data and typically process each frame independently. As a result, they cannot effectively exploit the spatiotemporal correlations present in point cloud sequences.We propose a unified geometry and attribute enhancement framework (DUGAE) for G-PCC compressed dynamic point clouds that explicitly exploits inter-frame spatiotemporal correlations in both geometry and attributes. First, a dynamic geometry enhancement network (DGE-Net) based on sparse convolution (SPConv) and feature-domain geometry motion compensation (GMC) aligns and aggregates spatiotemporal information. Then, a detail-aware k-nearest neighbors (DA-KNN) recoloring module maps the original attributes onto the enhanced geometry at the encoder side, improving mapping completeness and preserving attribute details. Finally, a dynamic attribute enhancement network (DAE-Net) with dedicated temporal feature extraction and feature-domain attribute motion compensation (AMC) refines attributes by modeling complex spatiotemporal correlations. On seven dynamic point clouds from the 8iVFB v2, Owlii, and MVUB datasets, DUGAE significantly enhanced the performance of the latest G-PCC geometry-based solid content test model (GeS-TM v10). For geometry (D1), it achieved an average BD-PSNR gain of 11.03 dB and a 93.95\% BD-bitrate reduction. For the luma component, it achieved a 4.23 dB BD-PSNR gain with a 66.61\% BD-bitrate reduction. DUGAE also improved perceptual quality (as measured by PCQM) and outperformed V-PCC. Our source code will be released on GitHub at: \url{https://github.com/yuanhui0325/DUGAE}
\end{abstract}

\begin{IEEEkeywords}
Dynamic point cloud compression, dynamic geometry enhancement, dynamic attribute enhancement, spatiotemporal correlations, motion compensation, sparse convolution.
\end{IEEEkeywords}

\section{Introduction}
\IEEEPARstart D{ynamic} point clouds represent objects or scenes as time-varying sets of 3D points with associated attributes (e.g., color and reflectance). They are widely used in autonomous driving~\cite{ref1}, virtual reality~\cite{ref2},and cultural heritage preservation~\cite{ref3}. However, dynamic point clouds are highly data-intensive, posing major challenges for storage, transmission, and real-time processing.

To reduce the bitrate of dynamic point clouds, the Moving Picture Experts Group (MPEG) has proposed two standards: geometry-based point cloud compression (G-PCC)~\cite{ref4}, which encodes point clouds in 3D space and is particularly well suited to sparse point clouds, and video-based point cloud compression (V-PCC)~\cite{ref5}, which projects point clouds into 2D patches, packs them into geometry and attribute images, and compresses them using existing video codecs, often providing advantages for dense point clouds. In recent years, to further improve compression efficiency for solid and dynamic point clouds, MPEG developed a new branch of G-PCC, called geometry-based solid content test model (GeS-TM)~\cite{ref6}. In parallel, learning-based point cloud compression methods~\cite{ref7,ref8,ref9,ref10,ref11,ref30,ref31} have rapidly emerged. By exploiting large-scale data and deep networks, these methods can achieve better rate--distortion (R--D) performance than traditional codecs. However, they typically require GPU-accelerated inference and suffer from reproducibility issues across heterogeneous hardware and software platforms~\cite{ref12,ref13}. In contrast, G-PCC and V-PCC offer stable and reproducible decoding across devices, but their compression performance can lag behind recent learned approaches for some content types and operating points.

To retain the interoperability and reproducibility of traditional codecs while exploiting the strong artifact-removal capability of deep models, many post-decoding quality enhancement methods~\cite{ref14,ref15,ref16,ref17,ref18,ref19,ref20,ref21,ref22,ref23,ref24,ref25,ref26} have been proposed for compressed point clouds. For geometry enhancement, DGPP~\cite{ref14} is one of the earliest methods for geometry enhancement of G-PCC-compressed point clouds. Later approaches based on sparse tensors and sparse convolution have been developed to enhance large-scale point clouds more efficiently. For attribute enhancement, MS-GAT \cite{ref17} and GQE-Net \cite{ref18} typically partition decoded point clouds into local blocks and apply graph-based operators to mitigate compression-induced distortions. G-PCC++ \cite{ref15} first densifies and enhances the decoded geometry, then maps the decoded attributes onto the enhanced geometry and refines them. UGAE \cite{ref23} is another unified enhancement method for G-PCC compressed static point clouds. Compared with G-PCC++, which enhances geometry and then refines attributes at the decoder side, UGAE restores an enhanced geometry at the encoder and recolors it using the original attributes to form an intermediate point cloud for subsequent attribute compression. At the decoder, the same enhanced geometry is reconstructed to assist attribute decoding, and an additional refinement step further improves the decoded attributes to produce the final enhanced point cloud. However, most existing methods are designed for static point clouds. Moreover, current dynamic point cloud quality enhancement methods typically focus on either geometry~\cite{ref18,ref24} or attributes~\cite{ref19,ref27} rather than addressing both jointly. 

To address these limitations, we propose a unified quality enhancement framework for compressed dynamic point clouds. Rather than optimizing the joint objective directly, which is challenging because geometry and attributes are strongly interdependent, we decompose the task into three subproblems: dynamic geometry enhancement, attribute recoloring, and dynamic attribute enhancement. This decomposition enables dedicated optimization of each component and improves the overall enhancement performance. Based on this framework, we develop DUGAE, a unified geometry and attribute enhancement method for dynamic point clouds. To the best of our knowledge, DUGAE is the first unified enhancement framework specifically designed for compressed dynamic point cloud sequences. DUGAE consists of three components: a dynamic geometry enhancement network (DGE-Net), a detail-aware k-nearest neighbors (DA-KNN) recoloring module, and a dynamic attribute enhancement network (DAE-Net), corresponding to the three subproblems, respectively. In DGE-Net, we use a sparse convolution (SPConv)~\cite{ref28}-based U-Net~\cite{ref29} for geometry feature extraction to obtain multi-scale spatiotemporal features from the compressed geometry. Moreover, we design a feature-domain geometry motion compensation module (GMC), which uses generalized sparse convolution (GSConv) to align the geometries of adjacent frames. To exploit the enhanced geometry more effectively for attribute enhancement, DUGAE applies DA-KNN recoloring at the encoder side to map the original attributes onto the enhanced geometry, followed by attribute compression, so that richer attribute details can be preserved. DAE-Net further enhances attributes at the decoder side. Since attribute information is more complex than geometry information, using a single feature extractor for both temporal and spatial features makes it difficult to distinguish their characteristics. We therefore introduce a dedicated attribute temporal feature extractor (ATFE) to learn temporal attribute features. To align spatiotemporal attribute features, we further propose an attribute motion compensation module (AMC) composed of GSConv and a dense module (DM) designed for the complexity of attribute information.

The main contributions of this work are summarized as follows:
\begin{itemize}
\item We analyze the limitations of existing unified enhancement methods and propose a unified joint enhancement framework that decomposes the original problem into three subproblems: dynamic geometry enhancement, attribute recoloring, and dynamic attribute enhancement.
\item Based on this framework, we propose DUGAE, a unified geometry and attribute enhancement method for G-PCC compressed dynamic point clouds. DUGAE consists of three components: DGE-Net for dynamic geometry enhancement, DA-KNN for attribute recoloring, and DAE-Net for dynamic attribute enhancement.
\item In DGE-Net, we introduce a GSConv-based geometry motion compensation module (GMC) that aligns geometry features across adjacent frames via kernel shifting, enabling feature-domain modeling of spatiotemporal correlations.
\item In DAE-Net, we introduce an attribute temporal feature extractor (ATFE) that combines SPConv layers and dense modules to efficiently extract temporal attribute features. ATFE is trained end to end together with an attribute feature extractor (AFE), enabling DAE-Net to distinguish between temporal and spatial information.
\item We propose an attribute motion compensation module (AMC) that uses GSConv to align spatial and temporal attribute features in the feature domain and uses a dense module to enhance the aligned features across channels.
\end{itemize}

The rest of this paper is structured as follows. Section~\ref{sec:related} surveys related work on point cloud geometry enhancement and attribute enhancement. Section~\ref{sec:problem} formulates the quality enhancement problem for dynamic point clouds. Section~\ref{sec:method} details the proposed DUGAE framework, including the overall pipeline and the design of each module. Section~\ref{sec:results} reports experimental results and ablation studies to validate the effectiveness of our method. Section~\ref{sec:conclusion} concludes the paper.

\section{Related Work}
\label{sec:related}
\subsection{Geometry Enhancement}
Early works mainly focused on upsampling sparse, noisy, and low-resolution point cloud geometry. PU-Net is one of the earliest deep learning-based upsampling methods, which uses PointNet++~\cite{ref36} to extract multi-scale features and multi-branch MLPs to expand these features. PU-GAN~\cite{ref63} further incorporates a generative adversarial network (GAN) into the upsampling process. Liu \emph{et al.}~\cite{ref64} further introduced coarse feature expansion and geometric refinement to restore a dense point cloud in a coarse-to-fine manner. Then, they~\cite{ref65} introduced PUFA-GAN, a frequency-aware upsampling network capable of producing dense points on the target surface while simultaneously attenuating high-frequency noise. Later, they proposed PU-Mask~\cite{ref66}, which introduces a virtual masking mechanism to guide upsampling towards locally sparse regions. Subsequently, they~\cite{ref67} proposed PU-GSM, which learns an implicit point distribution on the underlying surface of a sparse point cloud from a global self-similarity perspective and then improves the upsampling quality using a gradient-aware dual-domain refiner and latent vector matching. PU-Dense~\cite{ref68} uses multi-scale SPConv layers together with a binary cross-entropy (BCE) loss, enabling efficient upsampling of point clouds containing millions of points.

For compressed point clouds, Fan \emph{et al.}~\cite{ref14} proposed DGPP, which uses 3D convolutions combined with a coarse-to-fine multi-scale probabilistic occupancy prediction strategy. Liu \emph{et al.}~\cite{ref16} proposed GRNet, which includes two enhancement strategies (coordinate expansion and coordinate refinement) to handle different types of point clouds. However, the above methods focus only on geometry enhancement for static point clouds, while work on dynamic point clouds remains limited. Liu \emph{et al.}~\cite{ref24} proposed AuxGR, which exploits information from the previous frame to construct an auxiliary bitstream for geometry enhancement, where the auxiliary bitstream efficiently encapsulates spatiotemporal information. Building on GRNet, Zhang \emph{et al.}~\cite{ref26} proposed GeoQE, which introduces a spatiotemporal feature fusion module that leverages correlations within the current frame and adjacent frames to further enhance point cloud geometry.

\subsection{Attribute Enhancement}
For G-PCC compressed point cloud attributes, Sheng \emph{et al.}~\cite{ref17} proposed MS-GAT, which adopts a multi-scale graph attention network that builds graphs from geometry coordinates and incorporates multi-scale neighborhoods together with quantization-step information to adaptively remove attribute artifacts. Xing \emph{et al.}~\cite{ref18} introduced a graph-based quality enhancement network, GQE-Net, which uses normal vectors and geometry distances as auxiliary information. Guo \emph{et al.}~\cite{ref25} proposed PCE-GAN, a point cloud quality enhancement GAN based on optimal transport theory, which uses dynamic graph attention for local feature extraction and a Transformer-style global correlation module to jointly optimize data fidelity and perceptual quality. These graph-based methods require partitioning the point cloud into small patches for enhancement, leading to high inference costs. SPConv-based approaches alleviate this issue. Zhang \emph{et al.}~\cite{ref69} proposed two attribute enhancement schemes based on multi-scale feature aggregation for adaptive offsets and a lightweight learnable bilateral filter, providing a practical trade-off between performance and complexity. Subsequently, they~\cite{ref70} proposed ARNet, which uses a two-branch SPConv-based network to generate multiple most-probable sample offsets and linearly combines them to reduce attribute artifacts. Guo \emph{et al.}~\cite{ref71} proposed a high-efficiency Wiener filter enhancement method that improves point cloud reconstruction quality through coefficient inheritance, luma-based point classification, and Morton code-based fast nearest-neighbor search. For V-PCC compressed point cloud attributes, Gao \emph{et al.}~\cite{ref21} proposed OCARNet, which feeds occupancy maps as priors and uses multi-level feature fusion with masked loss to suppress attribute distortions in occupied regions. Xing \emph{et al.}~\cite{ref22} proposed SSIU-Net, which maps distorted 3D point cloud patches to 2D images, enhances them with a lightweight U-Net architecture, and then maps the enhanced 2D results back to 3D, effectively improving attribute quality.

However, these methods target attribute enhancement for static point clouds. For dynamic point cloud attribute quality enhancement, Wei \emph{et al.}~\cite{ref72} proposed DCFA, which improves reconstruction quality through level-based and point-based adaptive quantization and Wiener filter-based refinement-level quality enhancement. Liu \emph{et al.}~\cite{ref19} proposed DAE-MP, which enhances attributes via fast inter-frame motion prediction, motion consistency constraints, and frequency-component separation and fusion. Guo \emph{et al.}~\cite{ref27} proposed STQE, which exploits recoloring-based motion compensation, channel-aware temporal attention, and Gaussian-guided neighborhood aggregation to jointly model spatiotemporal correlations and thus improve the attribute quality.

In terms of unified geometry and attribute enhancement, Zhang \emph{et al.}~\cite{ref15} proposed G-PCC++, which considers geometry and attribute distortions in a unified way, but it only operates at the decoder side. It first enhances the lossy geometry, then interpolates and enhances the attributes, which prevents fully exploiting the benefits of enhanced geometry for attribute enhancement. We~\cite{ref23} proposed UGAE, a unified geometry and attribute enhancement framework that integrates enhanced geometry into the attribute compression and enhancement pipeline. This integration mitigates distortions in both domains and significantly improves the overall quality of compressed point clouds.

In general, current quality enhancement methods for dynamic point clouds typically focus on either geometry or attributes, without fully exploiting the correlations between the two. Moreover, existing joint quality enhancement methods are designed only for static point clouds, and there is still no joint quality enhancement approach for dynamic point clouds, where the spatiotemporal correlations between adjacent frames remain underexploited.

\section{Problem Statement}
\label{sec:problem}
Let $\mathbf{P}=(\mathbf{G},\mathbf{A})$ be an original point cloud, where
$\mathbf{G}\in\mathbb{R}^{N\times 3}$ contains the 3D geometry coordinates and
$\mathbf{A}\in\mathbb{R}^{N\times 3}$ contains the associated attribute information (RGB color). Here, $N$ is the number of points in the point cloud, and the $i$-th rows of $\mathbf{G}$ and $\mathbf{A}$ correspond to the geometry and attributes of the $i$-th point, respectively. After compression and decoding, we obtain the reconstructed point cloud $\mathbf{\bar P}=(\mathbf{\bar G},\mathbf{\bar A})$. We then enhance $\mathbf{\bar G}$ and $\mathbf{\bar A}$ to obtain the enhanced geometry $\hat{\mathbf{G}}$ and enhanced attributes $\hat{\mathbf{A}}$, respectively. The static point cloud quality enhancement problem can be formulated as the multi-objective optimization problem 
\begin{equation}
\min_{\hat{\mathbf{G}},\,\hat{\mathbf{A}}}\; (d_G(\mathbf{G},\hat{\mathbf{G}}), d_A(\mathbf{A},\hat{\mathbf{A}})),
\label{eq:static_obj}
\end{equation}
where $d_G(\cdot)$ and $d_A(\cdot)$ denote the geometry and attribute distortion functions, respectively.

In G-PCC++, quality enhancement is applied at the decoder side. After obtaining the enhanced geometry $\hat{\mathbf{G}}$, the decoded attributes $\mathbf{\bar A}$ are mapped onto the enhanced geometry $\hat{\mathbf{G}}$ to obtain recolored attributes $\mathbf{A}'$, which are then further enhanced. The process can be written as 
\begin{equation}
\begin{aligned}
& \hat{\mathbf{G}} = E_{\text{G-PCC++}}^{G}(\mathbf{\bar G}), \\
& \mathbf{A}' = \text{Map}(\mathbf{\bar A},\hat{\mathbf{G}}), \\
& \hat{\mathbf{A}} = E_{\text{G-PCC++}}^{A}(\mathbf{A}'),
\end{aligned}
\label{eq:gpccpp_form}
\end{equation}
where $E_{\text{G-PCC++}}^{G}(\cdot)$ denotes the geometry quality enhancement module of G-PCC++, $E_{\text{G-PCC++}}^{A}(\cdot)$ denotes the attribute quality enhancement module of G-PCC++, and $\text{Map}(\cdot)$ denotes the point cloud recoloring operation. During training, the modules are learned by minimizing $d_G(\mathbf{G},\hat{\mathbf{G}}) + d_A(\mathbf{A},\hat{\mathbf{A}})$. As $\hat{\mathbf{G}}$ can differ substantially from $\mathbf{\bar G}$ in both point count and local structure, establishing reliable correspondences for recoloring is challenging, which can result in incomplete or inaccurate attribute mapping. Such mapping errors make recoloring sensitive to small perturbations in the decoded geometry and attributes, and the resulting errors can be further propagated, or even amplified, by the subsequent enhancement module. Consequently, attribute enhancement becomes less stable and more difficult to optimize reliably.

To improve the completeness of the mapping and mitigate this instability, we move recoloring to the encoder side by mapping the original attributes $\mathbf{A}$ onto the enhanced geometry $\hat{\mathbf{G}}$ to obtain recolored attributes $\tilde{\mathbf{A}}$, forming $\tilde{\mathbf{P}}=(\hat{\mathbf{G}},\tilde{\mathbf{A}})$. During training, we need original point cloud $\mathbf{G}$ to generate occupancy labels. During inference, we only need the number of voxels $K$ in $\mathbf{G}$ and select the Top-$K$ voxels according to occupancy probabilities to form the enhanced geometry. Because the geometry enhancement module is deterministic, both the encoder and the decoder can obtain the same enhanced geometry $\hat{\mathbf{G}}$. After lossy attribute compression of $\tilde{\mathbf{A}}$, we obtain $\ddot{\mathbf{A}}$, which is then enhanced to $\hat{\mathbf{A}}$. The final enhanced point cloud is denoted by $\hat{\mathbf{P}}=(\hat{\mathbf{G}},\hat{\mathbf{A}})$. The optimization problem can then be reformulated as
{\setlength{\jot}{2pt}
\begin{equation}
\begin{aligned}
\min_{\hat{\mathbf{G}},\,\hat{\mathbf{A}}}\;& (d_G(\mathbf{G}, \hat{\mathbf{G}}), d_A(\mathbf{A}, \hat{\mathbf{A}})) \\
\text{s.t. }\;& \hat{\mathbf{G}} = E_{\text{static}}^{G}(\mathbf{\bar G}), \\
& \tilde{\mathbf{A}} = \text{Map}(\mathbf{A},\hat{\mathbf{G}}), \\
& \ddot{\mathbf{A}} = \text{COD}_{\text{G-PCC}}(\tilde{\mathbf{A}}), \\
& \hat{\mathbf{A}} = E_{\text{static}}^{A}(\ddot{\mathbf{A}}),
\end{aligned}
\label{eq:static_reform}
\end{equation}}
where $E_{\text{static}}^{G}(\cdot)$ denotes a geometry quality enhancement method for static point clouds, which is trained by BCE loss. $E_{\text{static}}^{A}(\cdot)$ denotes an attribute quality enhancement method for static point clouds, and $\text{COD}_{\text{G-PCC}}$ denotes the G-PCC attribute codec. However, (\ref{eq:static_reform}) exploits only information within the current frame.
\begin{figure}[t]
\centering
\includegraphics[width=\linewidth]{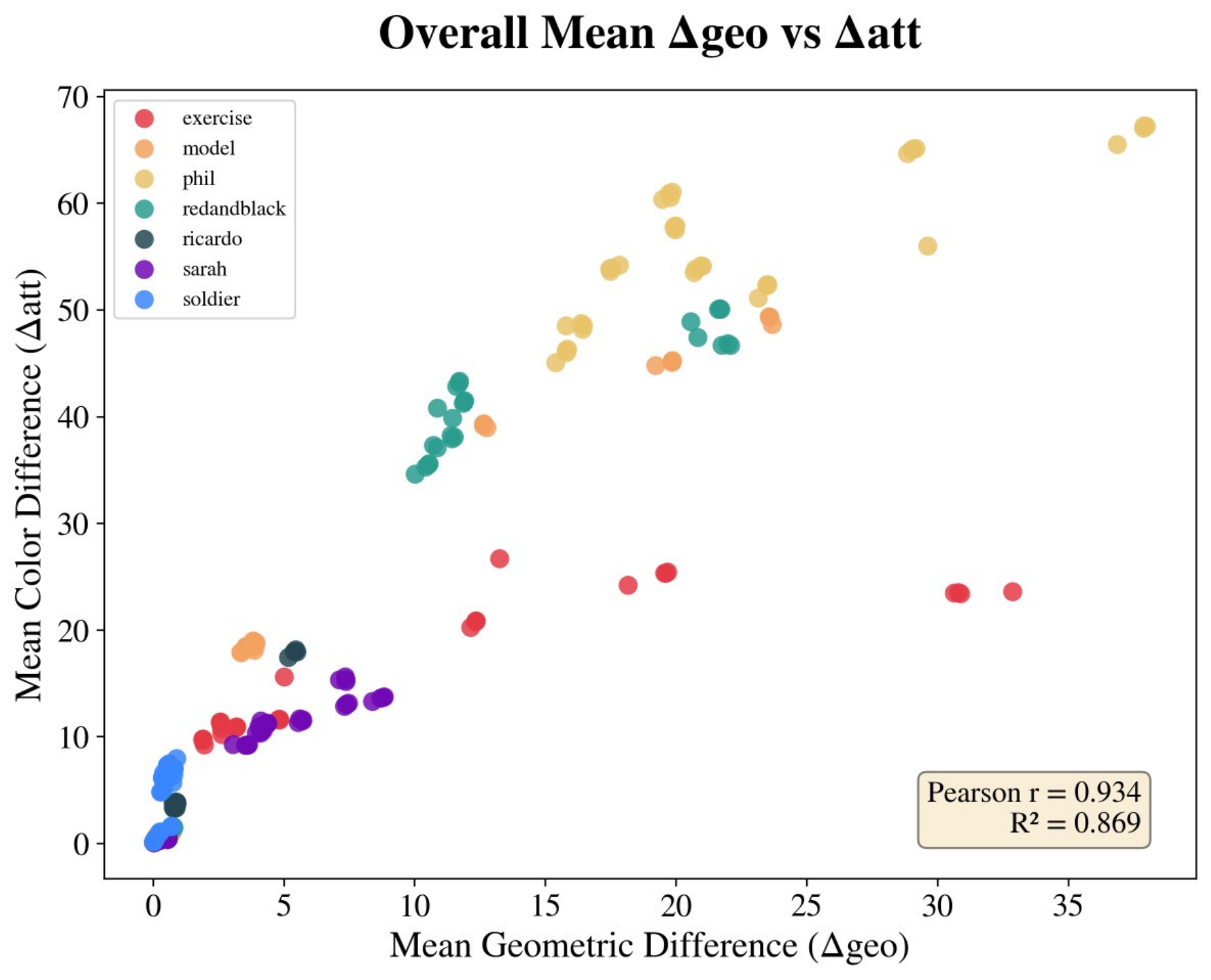}
\caption{Relationship between geometry error and color error.}
\label{fig:geo_color}
\end{figure}
For a dynamic point cloud sequence, let the current frame be $\mathbf{P}_t=(\mathbf{G}_t,\mathbf{A}_t)$. In GeS-TM, the previously decoded frame $\mathbf{\bar P}_{t-1}=(\mathbf{\bar G}_{t-1},\mathbf{\bar A}_{t-1})$ is available for prediction and is typically highly correlated with $\mathbf{P}_{t}$. To better exploit this temporal correlation, we extend (\ref{eq:static_reform}) by including information from the previously decoded frame in both geometry and attribute processing as follows: 
\begin{equation}
\begin{aligned}
\min_{\hat{\mathbf{G}}_t,\,\hat{\mathbf{A}}_t}\; & (d_G(\mathbf{G}_t, \hat{\mathbf{G}}_t) , d_A(\mathbf{A}_t, \hat{\mathbf{A}}_t)) \\
\text{s.t. }\;& \hat{\mathbf{G}}_t = E_{\text{dynamic}}^{G}\!(F_G(\mathbf{\bar G}_t,\mathbf{\bar G}_{t-1})), \\
& \tilde{\mathbf{A}}_t = \text{Map}(\mathbf{A}_t,\hat{\mathbf{G}}_t), \\
& \ddot{\mathbf{A}}_t = \text{COD}_{\text{GeS-TM}}\!(\tilde{\mathbf{A}}_t,\tilde{\mathbf{A}}_{t-1}), \\
& \hat{\mathbf{A}}_t = E_{\text{dynamic}}^{A}\!(F_A(\ddot{\mathbf{A}}_t,\ddot{\mathbf{A}}_{t-1})),
\end{aligned}
\label{eq:dynamic_form}
\end{equation}
where $E_{\text{dynamic}}^{G}(\cdot)$ and $E_{\text{dynamic}}^{A}(\cdot)$ denote the geometry and attribute quality enhancement modules for dynamic point clouds, respectively; $F_G(\cdot)$ fuses information from $\mathbf{\bar G}_t$ and $\mathbf{\bar G}_{t-1}$; $F_A(\cdot)$ fuses information from $\ddot{\mathbf{A}}_t$ and $\ddot{\mathbf{A}}_{t-1}$; and $\text{COD}_{\text{GeS-TM}}$ denotes the GeS-TM codec. Before attribute compression, we recolor the original attributes $\mathbf{A}_t$ onto the enhanced geometry $\hat{\mathbf{G}}_t$. As the enhanced geometry faithfully preserves the geometric structure and neighborhood relationships of the original point cloud and and has an almost identical number of points, the completeness of the mapping is improved. Moreover, since $\tilde{\mathbf{A}}_t$ retains a large amount of detail information, the subsequent enhancement stage becomes less sensitive to small compression-induced variations. 

\begin{figure*}[t]
\centering
\includegraphics[width=0.8\linewidth]{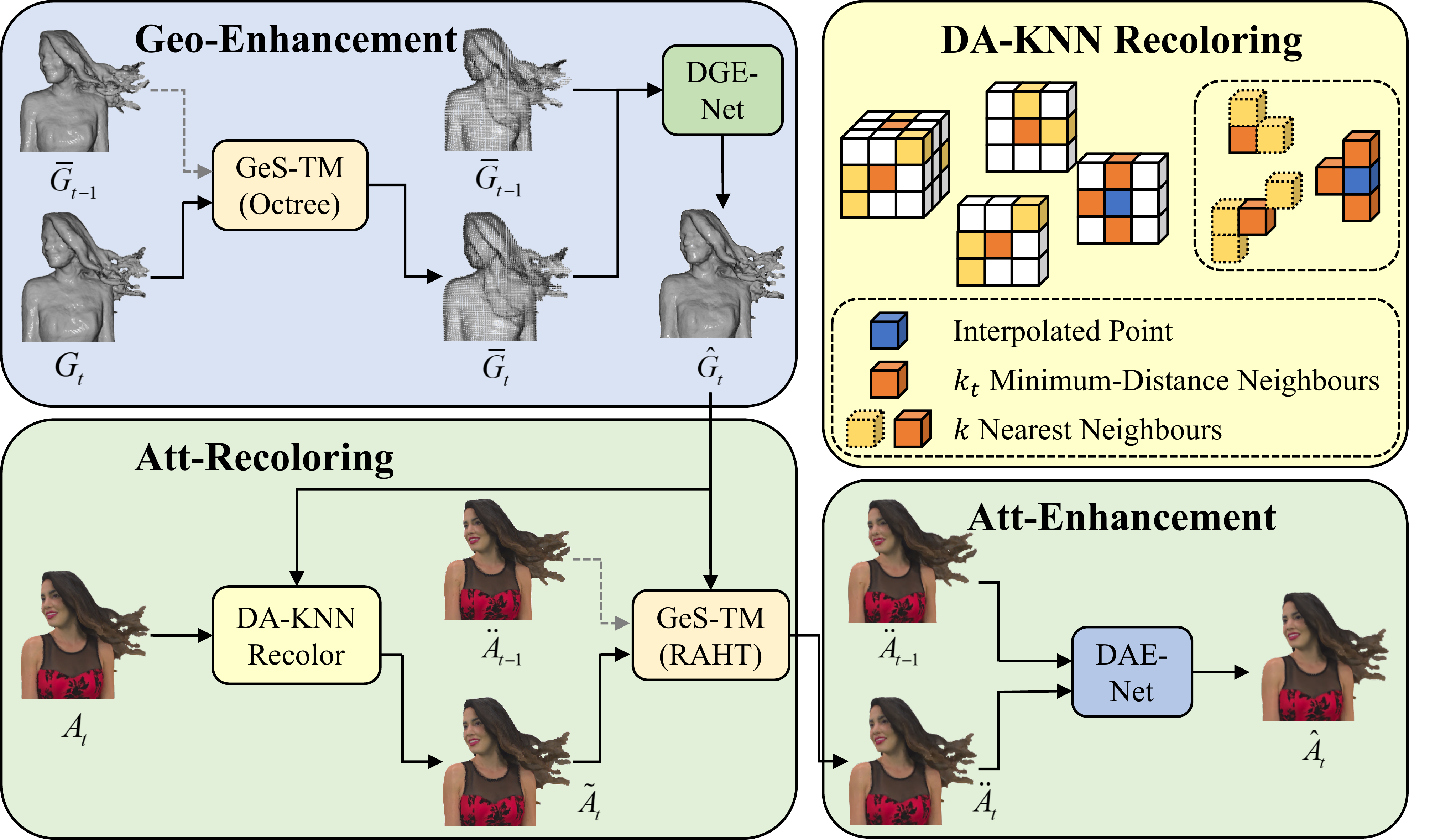}
\caption{Overall architecture of DUGAE for G-PCC compressed dynamic point clouds.}
\label{fig:framework}
\end{figure*}

However, $\mathbf{G}_t$ and $\hat{\mathbf{G}}_t$, as well as $\mathbf{A}_t$ and $\hat{\mathbf{A}}_t$, are defined on different point sets; thus their pointwise differences cannot be computed directly. To train the geometry enhancement module, we voxelize $\mathbf{G}_t$ to obtain ground-truth occupancy labels and use a BCE loss to encourage the voxel occupancy probabilities predicted by DGE-Net to match those of $\mathbf{G}_t$, which reduces the discrepancy between $\mathbf{G}_t$ and $\hat{\mathbf{G}}_t$. To train the attribute enhancement module, we measure attribute distortion on the common support $\hat{\mathbf{G}}_t$ using the squared error $d_A(\mathbf{A}_t, \hat{\mathbf{A}}_t)=\lVert \mathbf{A}_t-\hat{\mathbf{A}}_t \rVert_2^2$.

Using the inequality 
\begin{equation}
\begin{aligned}
\lVert \mathbf{A}_t-\hat{\mathbf{A}}_t \rVert_2^2 \\
&=\lVert (\mathbf{A}_t-\tilde{\mathbf{A}}_t) + (\tilde{\mathbf{A}}_t-\hat{\mathbf{A}}_t) \rVert_2^2 \\
&\le 2\lVert \mathbf{A}_t-\tilde{\mathbf{A}}_t \rVert_2^2
   +2\lVert \tilde{\mathbf{A}}_t-\hat{\mathbf{A}}_t \rVert_2^2,
\end{aligned}
\label{eq:upper_bound}
\end{equation}
the overall attribute discrepancy can be upper-bounded by twice the sum of a mapping error term $\lVert \mathbf{A}_t-\tilde{\mathbf{A}}_t\rVert_2^2$ and an enhancement error term $\lVert \tilde{\mathbf{A}}_t-\hat{\mathbf{A}}_t\rVert_2^2$. Accordingly, we use this upper bound for training by jointly reducing the mapping error and the enhancement error. Since $\tilde{\mathbf{A}}_t$ and $\hat{\mathbf{A}}_t$ are both defined on the same enhanced geometry $\hat{\mathbf{G}}_t$, the enhancement error can be computed directly without additional correspondence search. Moreover, the quality of $\tilde{\mathbf{A}}_t$ is largely determined by the quality of $\hat{\mathbf{G}}_t$: assuming $\mathrm{Map}(\mathbf{A}_t, \mathbf{G}_t)=\mathbf{A}_t$, when $\hat{\mathbf{G}}_t=\mathbf{G}_t$ we have $\tilde{\mathbf{A}}_t=\mathbf{A}_t$ and thus $\lVert \mathbf{A}_t-\tilde{\mathbf{A}}_t\rVert_2^2=0$. As the geometry distortion increases, the attribute error $\lVert \mathbf{A}_t-\tilde{\mathbf{A}}_t\rVert_2^2$ typically increases and may approach a worst-case bound for bounded attributes. Therefore, we assume that $d_A(\mathbf{A}_t,\tilde{\mathbf{A}}_t)$ is positively correlated with $d_G(\mathbf{G}_t,\hat{\mathbf{G}}_t)$.

To verify this hypothesis, we analyze the relationship between geometry and attribute errors in Fig.~\ref{fig:geo_color}. Because the point positions differ between point sets, we first locate, in the original point cloud $\mathbf{G}_t$, the nearest point
${p_t}^i = ({g_t}^i,{a_t}^i)$ to the current point
${\widetilde{p}}_t^i = (\hat{g}_t^i,\widetilde{a}_t^i)$, then compute the geometry and attribute error magnitudes as
$\Delta_{\text{geo}} = \|g_t^i-\hat{g}_t^i\|$ and
$\Delta_{\text{att}} = \|a_t^i-\widetilde{a}_t^i\|$, respectively.

The scatter points in Fig.~1 shows an overall increasing trend, which indicates a positive correlation between $\Delta_{\text{geo}}$ and $\Delta_{\text{att}}$. As attributes vary across point clouds and nearest-neighbor matching is imperfect, some samples appear as outliers. Nevertheless, the overall trend indicates that smaller geometry error is typically associated with smaller color error. We therefore model their relationship with an empirical linear fit,
\begin{equation}
\Delta_{\text{att}} = k\,\Delta_{\text{geo}}.
\label{eq:linear_fit}
\end{equation}

Based on the above analysis, we approximate the original optimization objective by a weighted-sum and decouple geometry and attribute processing as
\begin{equation}
\begin{aligned}
\min_{\hat{\mathbf{G}}_t,\,\hat{\mathbf{A}}_t}\;& C\cdot d_G(\mathbf{G}_t, \hat{\mathbf{G}}_t) + d_A(\tilde{\mathbf{A}}_t, \hat{\mathbf{A}}_t) \\
\text{s.t. }\;& \hat{\mathbf{G}}_t = E_{\text{dynamic}}^{G}\!(F_G(\mathbf{\bar G}_t,\mathbf{\bar G}_{t-1})), \\
& \tilde{\mathbf{A}}_t = \text{Map}(\mathbf{A}_t,\hat{\mathbf{G}}_t), \\
& \ddot{\mathbf{A}}_t = \text{COD}_{\text{GeS-TM}}\!(\tilde{\mathbf{A}}_t,\tilde{\mathbf{A}}_{t-1}), \\
& \hat{\mathbf{A}}_t = E_{\text{dynamic}}^{A}\!(F_A(\ddot{\mathbf{A}}_t,\ddot{\mathbf{A}}_{t-1})),
\end{aligned}
\label{eq:final_obj}
\end{equation}
where $C>0$ is a weighting constant. In practice, we train the geometry and attribute enhancement networks in a decoupled manner via the following two subproblems.
\begin{equation}
\begin{aligned}
\min_{\hat{\mathbf{G}}_t}\;& C\cdot d_G(\mathbf{G}_t, \hat{\mathbf{G}}_t) \\
=\;& \arg\min_{\text{DGE}}\;
d_G\!\Big(\mathbf{G}_t, \text{DGE}\!\big(F_G(\mathbf{\bar G}_t,\mathbf{\bar G}_{t-1})\big)\Big),
\end{aligned}
\label{eq:sub_geo}
\end{equation}
and 
\begin{equation}
\begin{aligned}
\min_{\hat{\mathbf{A}}_t}\;& d_A(\mathbf{A}_t, \hat{\mathbf{A}}_t) \\
=\;& \arg\min_{\mathrm{DAE}}\;
d_A\!\Big(\mathrm{DA\_KNN}(\mathbf{A}_t,\hat{\mathbf{G}}_t) , {}\\
&\qquad \mathrm{DAE}\!\big(F_A(\ddot{\mathbf{A}}_t,\ddot{\mathbf{A}}_{t-1})\big)\Big),
\end{aligned}
\label{eq:sub_att}
\end{equation}

where $\text{DGE}(\cdot)$ is the network implementation of the dynamic geometry quality enhancement method $E_{\text{dynamic}}^{G}(\cdot)$, $\text{DAE}(\cdot)$ is the network implementation of the dynamic attribute quality enhancement method $E_{\text{dynamic}}^{A}(\cdot)$, $\text{DA\_KNN}(\cdot)$ is the implementation of the point cloud attribute recoloring operator $\text{Map}(\cdot)$.

\begin{figure*}[t]
\centering
\includegraphics[width=\linewidth]{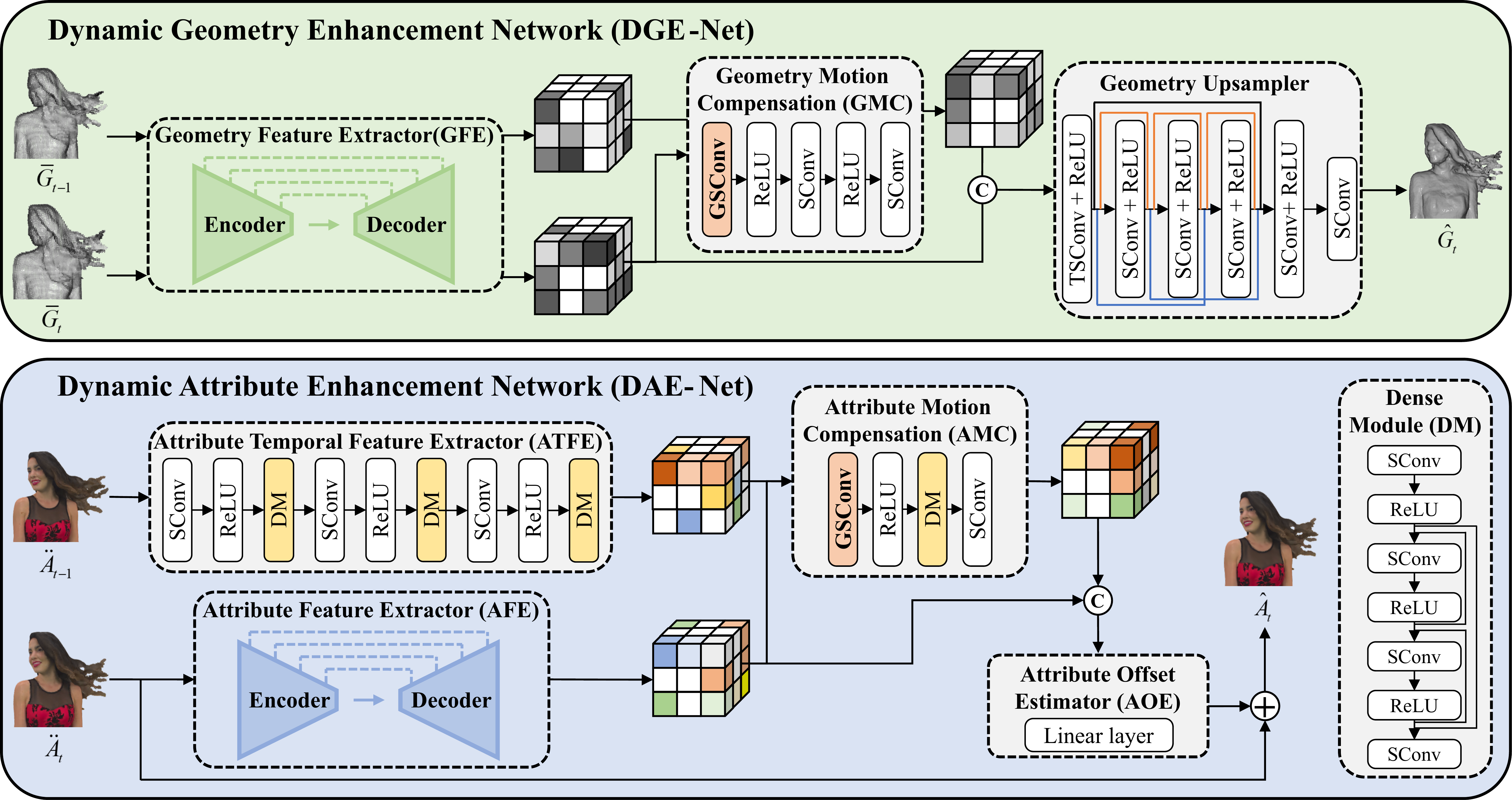}
\caption{DGE-Net and DAE-Net architectures.}
\label{fig:network}
\end{figure*}

\section{Proposed Method}
\label{sec:method}
\subsection{Overall Framework}
The overall architecture of DUGAE based on the GeS-TM compression pipeline is illustrated in Fig.~\ref{fig:framework}. DUGAE mainly consists of three components: dynamic geometry enhancement, attribute recoloring, and dynamic attribute enhancement. At the encoder side, DGE-Net takes the reconstructed geometries $\mathbf{\bar G}_{t-1}$ and $\mathbf{\bar G}_t$ from GeS-TM as inputs and outputs the enhanced geometry $\hat{\mathbf{G}}_t$. For attribute recoloring, the enhanced geometry $\hat{\mathbf{G}}_t$ serves as the attribute carrier. The original attributes $\mathbf{A}_t$ are mapped onto $\hat{\mathbf{G}}_t$ using DA-KNN, yielding $\tilde{\mathbf{A}}_t$. Since $\hat{\mathbf{G}}_t$ preserves the structure and details of the original geometry $\mathbf{G}_t$ to a high degree, the recolored point cloud $\tilde{\mathbf{P}}_t=(\hat{\mathbf{G}}_t,\tilde{\mathbf{A}}_t)$ retains rich high-frequency details. As DGE-Net is deterministic, the same enhanced geometry $\hat{\mathbf{G}}_t$ can also be obtained at the decoder side. Lossy attribute compression is then applied to the recolored attributes $\tilde{\mathbf{A}}_t$, which trades a small increase in bitrate for a substantial gain in high-frequency attribute information. At the decoder side, the compressed recolored attributes $\ddot{\mathbf{A}}_{t-1}$ and $\ddot{\mathbf{A}}_t$ are fed into DAE-Net to further enhance $\ddot{\mathbf{A}}_t$, producing the enhanced attributes $\hat{\mathbf{A}}_t$. The final enhanced point cloud is denoted as $\hat{\mathbf{P}}_t=(\hat{\mathbf{G}}_t,\hat{\mathbf{A}}_t)$.

\subsection{Dynamic Geometry Enhancement}
Dynamic geometry enhancement is implemented by DGE-Net. As shown in the upper part of Fig.~\ref{fig:network}, DGE-Net takes the compressed geometries of the previous and current frames as inputs. GFE extracts temporal and spatial features. The temporal features are then aligned to the spatial features through GMC. The aligned features are concatenated and fed into the geometry upsampler, which outputs the enhanced geometry of the current frame.

\subsubsection{Geometry Feature Extractor}
The global structure and local details of a point cloud provide important cues for geometry enhancement. To extract multi-level geometry features, the geometry feature extractor (GFE) in DGE-Net adopts a U-Net architecture constructed by stacking nine Point Transformer v3 (PT)~\cite{ref73} blocks, which is consistent with UGAE. The encoder contains five PT blocks. Through four pooling operations, the voxel occupancy features of the input point cloud are progressively downsampled to a 512-dimensional latent representation. At this stage, repeated pooling yields features with rich global information. To further capture multi-scale information, the decoder uses four skip connections and four unpooling operations, and outputs 32-dimensional multi-scale geometry features. This process can be written as
\begin{equation}
\begin{split}
(\mathbf{G}^{\mathrm{init}},\mathbf{F}^{\mathrm{init}}) &= \mathrm{Init}(\mathbf{G},\mathbf{M}_{\mathrm{occ}}), \\
(\mathbf{G}_{E}^{1},\mathbf{F}_{E}^{1}) &= E_{1}(\mathbf{G}^{\mathrm{init}},\mathbf{F}^{\mathrm{init}}), \\
(\mathbf{G}_{E}^{s},\mathbf{F}_{E}^{s}) &= E_{s}\!\Big(\mathrm{Pool}(\mathbf{G}_{E}^{s-1},\mathbf{F}_{E}^{s-1})\Big),\quad s=2,3,4,5, \\
(\mathbf{G}_{D}^{5},\mathbf{F}_{D}^{5}) &= (\mathbf{G}_{E}^{5},\mathbf{F}_{E}^{5}), \\
(\mathbf{G}_{D}^{s},\mathbf{F}_{D}^{s}) &= D_{s}\!\Big(\mathrm{UnPool}(\mathrm{Concat}(\mathbf{G}_{D}^{s+1},\mathbf{F}_{D}^{s+1},\mathbf{G}_{E}^{s},\mathbf{F}_{E}^{s}))\Big), \\
&\hspace{8.8em} s=4,3,2,1.
\end{split}
\label{eq:gfe}
\end{equation}
where $\mathbf{M}_{\mathrm{occ}}$ denotes the occupancy labels of the voxelized point cloud and all its entries are equal to $1$.
$(\mathbf{G}_E^{s},\mathbf{F}_E^{s})$ represent the geometry coordinates and geometry features at encoder level $s$ (i.e., the outputs of the $s$-th encoder block), and $(\mathbf{G}_D^{s},\mathbf{F}_D^{s})$ represent the geometry coordinates and geometry features at decoder level $s$ (i.e., the outputs of the $s$-th decoder block). Here, $s$ denotes the layer index, where the encoder has five layers and the decoder has four layers. $\mathrm{Init}(\cdot)$ denotes the initialization operation that converts an unordered point cloud into an ordered point sequence~\cite{ref73}. $E_{1}$ denotes the first encoder layer, while $E_{s}$ and $D_{s}$ denote the encoder and decoder at the $s$-th layer, respectively. $\mathrm{Pool}(\cdot)$ and $\mathrm{UnPool}(\cdot)$ denote the pooling and unpooling operations, respectively.

In dynamic point clouds, the compressed lossy geometries $\mathbf{\bar G}_{t-1}$ and $\mathbf{\bar G}_t$ exhibit similar topology. However, $\mathbf{\bar G}_{t-1}$ is sparse and the correlations between points are relatively weak. To fully exploit its topological information and assist the enhancement of the current geometry $\mathbf{\bar G}_t$, we use GFE with skip connections and attention layers to extract multi-scale temporal features with a large receptive field.

\subsubsection{Geometry Motion Compensation}
Dynamic point clouds capture different states of the same object or scene at successive time instants, so adjacent frames typically exhibit strong spatiotemporal correlations. Temporal and spatial features are extracted from different point sets whose coordinates are not aligned, so they cannot be fused pointwise without first establishing correspondences. D-DPCC~\cite{ref74} uses $k$-nearest neighbors (KNN) search to align geometries, while AuxGR~\cite{ref24} uses KNN-based neighborhoods with cross-attention to capture motion. However, running KNN for every point is computationally expensive and can significantly increase runtime.

We propose geometry motion compensation (GMC) to align inter-frame geometry in the feature domain. The module consists of one generalized sparse convolution (GSConv), two sparse convolution (SPConv) layers, and two ReLU layers. GSConv maps features from an input coordinate set to a target coordinate set, allowing the input and output coordinates to differ and thus enabling efficient coordinate mapping in the feature domain. For an input sparse tensor $(\mathbf{G}_{\mathrm{in}},\mathbf{F}_{\mathrm{in}})$ and a target sparse tensor $(\mathbf{G}_{\mathrm{out}},\mathbf{F}_{\mathrm{out}})$, GSConv is defined as
\begin{equation}
\mathbf{f}^{\mathrm{out}}_{u}
=\sum_{k\in \mathcal{N}^{3}(u,\mathbf{G}_{\mathrm{in}})} \mathbf{W}_{k}\,\mathbf{f}^{\mathrm{in}}_{u+k},
\quad \forall\, u\in \mathbf{G}_{\mathrm{out}},
\label{eq:gsconv}
\end{equation}
where $\mathbf{f}^{\mathrm{out}}_{u}$ denotes the output feature at target coordinate $u$, and $\mathbf{f}^{\mathrm{in}}_{u+k}$ denotes the input feature at coordinate $u+k$. Here, $\mathcal{N}^{3}$ denotes the set of 3D convolution kernel offsets, and
$\mathcal{N}^{3}(u,\mathbf{G}_{\mathrm{in}})=\{\,k \mid u+k\in \mathbf{G}_{\mathrm{in}},\; k\in \mathcal{N}^{3}\,\}$. $\mathbf{W}_{k}$ is the convolution kernel weight corresponding to offset $k$. In this work, a $3\times 3\times 3$ convolution kernel is used to map the geometry features of the previous frame $\mathbf{F}^{G}_{t-1}$ to the geometry features of the current frame $\mathbf{F}^{G}_{t}$, injecting multi-scale temporal information from the previous frame into the current frame in the feature domain.

\subsubsection{Geometry Upsampler}
To enhance a dense point cloud from sparse point features, DGE-Net adopts a geometry upsampler formed by a transpose sparse convolution (TSConv) layer and a dense connection of four sparse convolution layers. TSConv first generates a large number of candidate points. The subsequent dense connection enhances the features in the channel domain to obtain more stable representations. The last sparse convolution layer outputs the voxel occupancy probabilities $\mathbf{M}_{p}$ for all candidate points. Given the number of points $N$ in the original point cloud, we select the Top-$N$ points in $\mathbf{M}_{p}$ as the enhanced geometry. To guarantee consistent results at the encoder and decoder, the TSConv layer is moved to the CPU during testing to avoid non-deterministic behavior introduced by the GPU.

\subsection{Attribute Recoloring}
During lossy compression in GeS-TM, the built-in recolor module can only map the rich original attributes $\mathbf{A}_t$ onto the sparse compressed geometry $\mathbf{\bar G}_t$. This operation introduces large distortions in attribute information, and subsequent compression further intensifies information loss. Benefiting from the deterministic property of DGE, we instead map the original attributes $\mathbf{A}_t$ onto the enhanced geometry $\hat{\mathbf{G}}_t$, whose point density and topological structure are close to those of the original geometry. The recolored attributes $\tilde{\mathbf{A}}_t$ are then encoded by lossy compression. In this way, richer attribute information can be reconstructed at the decoder side, which further benefits attribute enhancement.

\subsubsection{Recolor Algorithm}
Distance-weighted recoloring methods adjust the contribution of each reference point according to its distance, assigning smaller weights to distant points. Although this reduces the influence of far-away points, it inevitably causes attribute blurring. Existing methods~\cite{ref11,ref62} directly select the nearest point as the reference, which mitigates this problem; however, in many cases a point may have multiple equally close nearest neighbors. To fully exploit the enhanced geometry and achieve accurate recoloring, we introduce DA-KNN. As illustrated in Fig.~2, for a point $\hat{g}_i$ in the enhanced geometry $\hat{\mathbf{G}}_t$, we first select $q$ nearest neighbors from $\mathbf{G}_t$, and then choose $q_t$ minimum-distance neighbors among them to form a local neighborhood $\{(g_i^{l},a_i^{l})\}_{l=1}^{q_t}$, where $g_i^{l}$ and $a_i^{l}$ denote the geometry and attribute of the $l$-th neighbor, respectively. The recolored attribute $\widetilde{a}_i$ of the $i$-th point is computed as
\begin{equation}
\widetilde{a}_i=\frac{1}{q_t}\sum_{l=1}^{q_t} a_i^{l},
\quad \text{s.t. } d(\hat{g}_i,g_i^{l})=\min d(\hat{g}_i,\mathbf{G}_t),
\label{eq:da_knn}
\end{equation}
where $d(\cdot)$ denotes the Euclidean distance. The condition $d(\hat{g}_i,g_i^{l})=\min d(\hat{g}_i,\mathbf{G}_t)$ indicates that every point $g_i^{l}$ in the neighborhood has the same distance to $\hat{g}_i$, and this distance is the minimum distance from $\hat{g}_i$ to all points in the original geometry $\mathbf{G}_t$.

\subsection{Attribute Enhancement}
Even though the enhanced geometry provides a solid basis for attribute compression and helps preserve detail information, further enhancement of the attributes is still crucial, especially for high-frequency components. As shown in the lower part of Fig.~3, to enhance the compressed recolored attributes $\ddot{\mathbf{A}}_t$, we propose DAE-Net, which consists of ATFE, AFE, AMC, and an Attribute Offset Estimator (AOE). For feature extraction, DAE-Net is similar to DGE-Net in that AFE also adopts a U-Net architecture to extract multi-scale attribute features. The difference lies in the channel dimensions: the input and output channels change from 1 and 32 in DGE-Net to 3 and 64 in DAE-Net. Attributes are represented in the three-channel YUV space, and attribute information is much more complex than the geometry occupancy probability; thus, more channels are required to model it.

Unlike DGE-Net, which uses a single GFE to extract both spatial and temporal geometry features, DAE-Net must account for attribute enhancement operating on the enhanced geometry (which is finer than the compressed geometry) and that attribute information is more complex than geometry information. To effectively extract temporal information from the point cloud attributes $\ddot{\mathbf{A}}_{t-1}$, we design an additional ATFE composed of three sparse convolution (SConv) layers, three ReLU layers, and three dense modules (DMs). This design helps prevent the model from confusing temporal features with spatial features.

\subsubsection{Attribute Motion Compensation}
Similar to the motion compensation problem in geometry enhancement, the temporal attribute features are extracted based on the enhanced geometry $\hat{\mathbf{G}}_{t-1}$, which is not aligned with the current-frame geometry $\hat{\mathbf{G}}_t$. This misalignment prevents rich temporal information from being effectively used for attribute enhancement. To align temporal and spatial attribute features, GSConv is introduced again. AMC maps temporal features onto spatial features. A DM is inserted to fuse features across different channels and enhance them in the channel domain. After an SConv layer, we obtain temporally enhanced features aligned with the current frame.

\subsubsection{Attribute Offset Estimator}
Compared with binary occupancy prediction for geometry, directly regressing continuous attribute values is more challenging; therefore, predicting attribute offsets becomes an attractive alternative. After temporal and spatial attribute features are fused, a simple AOE consisting of a single linear layer maps the 128-dimensional fused feature at each point to a three-channel offset in the YUV color space. This offset is then added to the compressed recolored attributes $\ddot{\mathbf{A}}_t$ to obtain the final enhanced attributes $\hat{\mathbf{A}}_t$.

\subsection{Loss Function}
Dynamic geometry enhancement and dynamic attribute enhancement are trained in a decoupled manner, simplifying the global optimization into two subproblems. Given the current compressed point cloud $\mathbf{\bar P}_t=(\mathbf{\bar G}_t,\mathbf{\bar A}_t)$ and the previous compressed point cloud $\mathbf{\bar P}_{t-1}=(\mathbf{\bar G}_{t-1},\mathbf{\bar A}_{t-1})$, DGE-Net takes the compressed geometries $\mathbf{\bar G}_t$ and $\mathbf{\bar G}_{t-1}$ as inputs and outputs the enhanced geometry $\hat{\mathbf{G}}_t$. We train DGE-Net end-to-end with the BCE loss:
\begin{equation}
\ell_{\mathrm{BCE}}
=-\frac{1}{N_c}\sum_{i=1}^{N_c}\Big(o_i\log p_i + (1-o_i)\log(1-p_i)\Big),
\label{eq:bce}
\end{equation}
where $N_c$ is the number of candidate voxel locations output by the network. $o_i$ is the ground-truth voxel occupancy label, which equals $1$ for occupied voxels and $0$ for unoccupied voxels. $p_i$ is the predicted occupancy probability of the $i$-th candidate voxel locations.

\begin{table}[t]
  \centering
  \caption{Quantization Parameters of GeS-TM and V-PCC for All Dynamic Point Cloud Test Sequences.}
  \label{tab:qp}
  \renewcommand{\arraystretch}{1.2}
  \setlength{\tabcolsep}{4pt}
  \begin{tabular}{lccccc}
    \toprule
    Rate level     & R01   & R02   & R03  & R04  & R05 \\
    \midrule
    GeS-TM PQS     & 0.125 & 0.25  & 0.50 & 0.75 & 0.875 \\
    GeS-TM QP      & 51    & 46    & 40   & 34   & 28   \\
    V-PCC QP-G     & 32    & 28    & 24   & 20   & 16   \\
    V-PCC QP-A     & 42    & 37    & 32   & 27   & 22   \\
    \bottomrule
  \end{tabular}
\end{table}

For dynamic attribute enhancement, previous work~\cite{ref23} has shown that regions where the reconstructed attributes deviate significantly from the original attributes largely overlap with regions that contribute high MSE values. Guided by this observation, we train DAE-Net with a weighted MSE (W-MSE) loss to emphasize high-error regions. The decoded recolored attributes of the current frame $\ddot{\mathbf{A}}_t$ and those of the previous frame $\ddot{\mathbf{A}}_{t-1}$ are used as inputs to produce the enhanced attributes $\hat{\mathbf{A}}_t$. The loss is defined as
\begin{equation}
\ell_{\mathrm{W\text{-}MSE}}
=\frac{1}{N}\sum_{i=1}^{N} w_i \,\big\lVert \widetilde{a}_i-\hat{a}_i \big\rVert_2^{2},
\quad
w_i=
\begin{cases}
w_{\mathrm{high}}, & e_i > T,\\
w_{\mathrm{low}},  & \text{otherwise},
\end{cases}
\label{eq:wmse}
\end{equation}
where $\widetilde{a}_i$ is the $i$-th recolored attribute, $\hat{a}_i$ is the $i$-th enhanced attribute, and $w_i$ is the weight for each point. The threshold $T=\mathrm{Quantile}(\{e_i\}_{i=1}^{N},\tau)$ is set to the $\tau$-th quantile of all errors, where $e_i = |\widetilde{a}_i-\hat{a}_i|$ for $i=1,2,\ldots,N$ is the error between the recolored attribute $\widetilde{a}_i$ and the enhanced attribute $\hat{a}_i$ at the $i$-th point.

\section{Results and Analysis}
\label{sec:results}

We conducted extensive quantitative and qualitative experiments to thoroughly evaluate the performance of the proposed DUGAE under lossy compression of both geometry and attributes for dynamic point clouds. We first compare DUGAE with the geometry-based solid content test model (GeS-TM v10) and then with the V-PCC test model v25. We also compared against the unified enhancement method UGAE~\cite{ref23}, which is  for static point clouds and the deep learning--based dynamic point cloud geometry coding method DPCGC~\cite{ref75}. Moreover, we evaluated subjective visual quality and conducted ablation studies to assess the contribution of each component in DUGAE.

\begin{table*}[t]
\centering
\caption{Quantitative gains of DUGAE compared to GeS-TM v10.}
\label{tab:dugae_vs_ges_tm}
\renewcommand{\arraystretch}{1.15}
\setlength{\tabcolsep}{4pt}
\begin{tabular}{lccccc ccccc cc}
\toprule
\multirow{2}{*}{Point cloud} &
\multicolumn{5}{c}{BD-BR (\%)} &
\multicolumn{5}{c}{BD-PSNR (dB)} &
\multicolumn{2}{c}{$1$-PCQM} \\
\cmidrule(lr){2-6}\cmidrule(lr){7-11}\cmidrule(lr){12-13}
& D1 & D2 & Y & U & V & D1 & D2 & Y & U & V & BD-BR (\%) & BD-PCQM ($10^{-3}$) \\
\midrule
\textit{redandblack} & -92.59 & -89.68 & -63.60 & -55.15 & -57.84 & 11.30 & 10.30 & 3.85 & 2.00 & 3.56 & -78.15 & 16.68 \\
\textit{soldier}     & -95.19 & -92.83 & -65.29 & -41.69 & -55.97 & 11.51 & 10.61 & 3.36 & 1.64 & 2.37 & -84.70 & 16.90 \\
\textit{exercise}    & -95.59 & -93.61 & -65.20 & -63.72 & -66.20 & 12.78 & 11.58 & 3.30 & 1.48 & 2.14 & -79.24 &  8.47 \\
\textit{model}       & -94.54 & -91.71 & -66.06 & -67.34 & -70.98 & 12.34 & 10.67 & 4.35 & 1.50 & 3.02 & -80.90 & 13.32 \\
\textit{phil}        & -92.68 & -85.82 & -70.58 & -58.61 & -69.96 &  9.69 &  8.01 & 4.91 & 1.49 & 2.19 & -82.48 & 18.93 \\
\textit{ricardo}     & -94.23 & -86.36 & -68.07 & -64.71 & -72.72 &  9.87 &  7.81 & 4.69 & 2.43 & 2.98 & -74.72 &  5.83 \\
\textit{sarah}       & -92.86 & -85.95 & -67.47 & -68.98 & -69.72 &  9.74 &  8.05 & 5.11 & 3.41 & 3.73 & -78.06 &  7.56 \\
\midrule
\textbf{Average}         & \textbf{-93.95} & \textbf{-89.42} & \textbf{-66.61} & \textbf{-60.03} & \textbf{-66.20} &
\textbf{11.03} & \textbf{9.58} & \textbf{4.23} & \textbf{1.99} & \textbf{2.86} &
\textbf{-79.75} & \textbf{12.53} \\
\bottomrule
\end{tabular}
\end{table*}

\begin{table*}[t]
\centering
\caption{Quantitative gains of DUGAE compared to UGAE.}
\label{tab:dugae_vs_ugae}
\renewcommand{\arraystretch}{1.15}
\setlength{\tabcolsep}{4pt}
\begin{tabular}{lccccc ccccc cc}
\toprule
\multirow{2}{*}{Point cloud} &
\multicolumn{5}{c}{BD-BR (\%)} &
\multicolumn{5}{c}{BD-PSNR (dB)} &
\multicolumn{2}{c}{$1$-PCQM} \\
\cmidrule(lr){2-6}\cmidrule(lr){7-11}\cmidrule(lr){12-13}
& D1 & D2 & Y & U & V & D1 & D2 & Y & U & V & BD-BR (\%) & BD-PCQM ($10^{-3}$) \\
\midrule
\textit{redandblack} &  -8.58 &  0.37 & -4.30 & -11.72 &  -7.32 & 0.45 & 0.37 & 0.13 & 0.18 & 0.26 &  -1.32 &  0.03 \\
\textit{soldier}     & -10.76 & -4.45 & -1.76 & -34.66 & -25.41 & 0.51 & -4.45 & 0.04 & 0.66 & 0.51 &  -3.66 &  0.11 \\
\textit{exercise}    & -11.32 &  1.79 & -1.82 & -16.15 & -12.81 & 0.58 & 1.79 & 0.03 & 0.08 & 0.12 &  -3.32 &  0.15 \\
\textit{model}       &  -8.15 &  1.47 & -2.69 & -19.20 & -11.72 & 0.35 & 1.47 & 0.08 & 0.26 & 0.20 & -10.38 &  0.66 \\
\textit{phil}        &  -4.02 &  2.21 & -2.85 &  -9.18 &  -8.39 & 0.15 & 2.21 & 0.09 & 0.15 & 0.16 &   1.49 & -0.06 \\
\textit{ricardo}     &  -5.40 &  0.19 & -1.60 & -12.60 & -17.09 & 0.16 & 0.19 & 0.03 & 0.12 & 0.22 &  -2.86 &  0.06 \\
\textit{sarah}       &  -4.96 & -0.88 & -3.03 &  -9.59 & -13.74 & 0.18 & -0.88 & 0.08 & 0.14 & 0.26 &  -3.96 &  0.25 \\
\midrule
\textbf{Average}         & \textbf{-7.60} & \textbf{0.10} & \textbf{-2.58} & \textbf{-16.16} & \textbf{-12.78} &
\textbf{0.34} & \textbf{-0.02} & \textbf{0.07} & \textbf{0.23} & \textbf{0.25} &
\textbf{-3.43} & \textbf{0.17} \\
\bottomrule
\end{tabular}
\end{table*}

\subsection{Datasets}
\subsubsection{Training Datasets}
We trained the proposed DUGAE using six dynamic point cloud sequences: \textit{Longdress}, \textit{Loot}, \textit{Basketballplayer}, \textit{Dancer}, \textit{Andrew}, and \textit{David}. \textit{Longdress} and \textit{Loot} are from the 8i Voxelized Full Bodies dataset (8iVFB v2)~\cite{ref76}. \textit{Basketballplayer} and \textit{Dancer} are from the Owlii Dynamic Human Textured Mesh Sequence dataset (Owlii)~\cite{ref77}. \textit{Andrew} and \textit{David} are from the Microsoft Voxelized Upper Bodies dataset (MVUB)~\cite{ref78}. Due to GPU memory constraints, used a KD-tree-based partitioning strategy to split each voxelized 3D point cloud into sub-point clouds, each containing at most 100{,}000 points.

The resulting sub-point cloud sequences were then compressed using GeS-TM v10 with the octree-RAHT configuration to construct the training data. All compression experiments were done under the Common Test Conditions (CTC)~\cite{ref79}. We used the first 32 frames of each sequence for training, yielding a total of 192 frames. Note that DGE-Net was trained on both the original geometries of the sub-point cloud sequences and their compressed counterparts, whereas DAE-Net was trained on the recolored sub-point cloud sequences together with the corresponding compressed attribute point clouds.

\subsubsection{Test Datasets}
We evaluated the performance of DUGAE on seven dynamic point cloud sequences: \textit{Redandblack}, \textit{Soldier}, \textit{Exercise}, \textit{Model}, \textit{Phil}, \textit{Ricardo}, and \textit{Sarah}. \textit{Redandblack} and \textit{Soldier} are from the 8iVFB v2 dataset~\cite{ref76}. \textit{Exercise} and \textit{Model} are from the Owlii dataset~\cite{ref77}, while \textit{Phil}, \textit{Ricardo}, and \textit{Sarah} are from the MVUB dataset~\cite{ref78}. Table~\ref{tab:qp}shows key compression parameters used during compression: the GeS-TM geometry quantization parameter Position Quantization Scale (GeS-TM PQS), the GeS-TM attribute quantization parameter Quantization Parameter (GeS-TM QP), the V-PCC geometry quantization parameter geometry QP (V-PCC QP-G), and the V-PCC attribute quantization parameter attribute QP (V-PCC QP-A). For testing, we selected the first 32 frames from each sequence, resulting in 224 frames in total.

\subsection{Implementation Details}
We implemented DUGAE using PyTorch~2.1.1 and MinkowskiEngine~0.5.4, and carried out all experiments on a workstation equipped with an Intel Xeon Gold~6148 CPU, 640~GB RAM, and eight Nvidia RTX~4090 GPUs. Both DGE-Net and DAE-Net were optimized with the AdamW optimizer for 100 epochs. The learning rate was set to 0.004, and the batch sizes for DGE-Net and DAE-Net were 1 and 3, respectively. For recoloring, we used DA-KNN with $k=8$ neighbors, following the configuration used in UGAE. The W-MSE loss function used the same hyperparameters as UGAE, with $w_{\mathrm{high}}=2$, $w_{\mathrm{low}}=0.5$, and $\sigma=0.4$.

To evaluate the performance of DUGAE, we used bits per input point (bpip) to measure bitrate. Geometry reconstruction quality was quantified using point-to-point (D1 PSNR) and point-to-plane (D2 PSNR) metrics. For attribute quality, we report Y-PSNR for the luma component and YUV-PSNR~\cite{ref80} with a Y:U:V weighting ratio of 14:1:1 to characterize overall attribute fidelity. To more comprehensively capture perceptual visual quality, we additionally used the PCQM~\cite{ref81} metric for objective assessment of colored point clouds. For detailed R--D analysis, we computed Bjøntegaard Delta (BD)~\cite{ref82} metrics\footnote{To compute the BD metrics, we used the code: \url{https://github.com/FAU-LMS/bjontegaard-matlab}}, including BD-Bitrate (BD-BR), BD-PSNR, and BD-PCQM, using Akima interpolation to fit R--D curves. Negative BD-BR values indicate bitrate savings achieved by the proposed method.
\begin{figure*}[t]
\centering
\includegraphics[width=\linewidth]{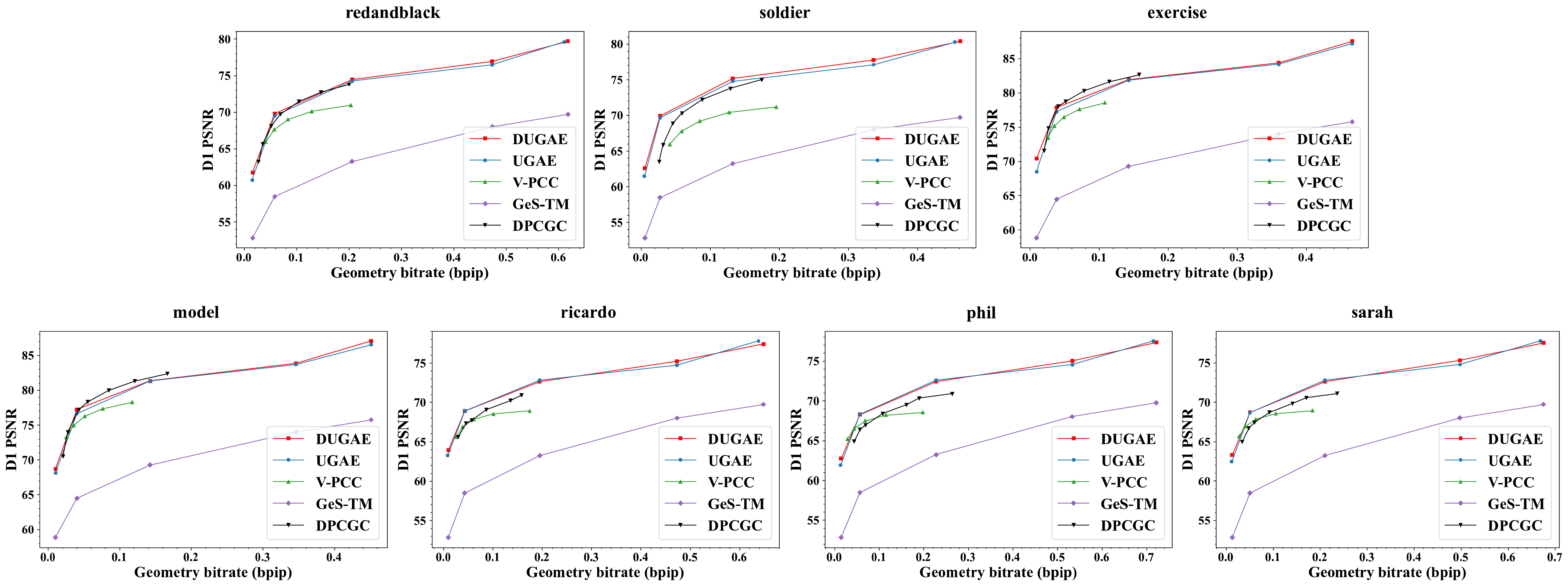}
\caption{D1-PSNR vs. geometry bitrate for DUGAE, UGAE, V-PCC, GeS-TM, and DPCGC.  
}
\label{fig:R-D_G}
\end{figure*}

\begin{figure*}[t]
\centering
\includegraphics[width=\linewidth]{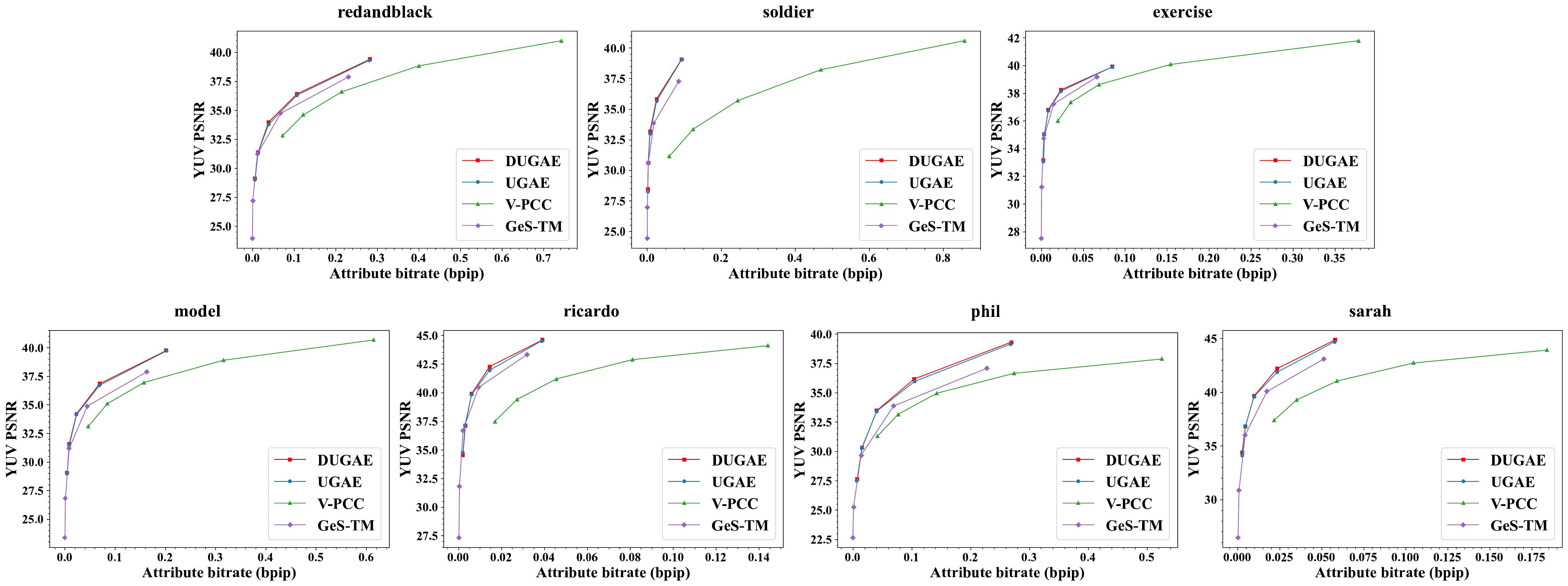}
\caption{YUV PSNR vs. attribute bitrate for DUGAE, UGAE, V-PCC, and Ge-S-TM.}
\label{fig:R-D_A}
\end{figure*}

\subsection{Objective Quality Evaluation}
Table~\ref{tab:dugae_vs_ges_tm} presents the objective quality evaluation results of DUGAE and GeS-TM on the seven test point cloud sequences. The results show that DUGAE achieved substantial improvements in both geometry and attribute quality under the same total bitrate. For geometry, DUGAE achieved average BD-PSNR gains of 11.03~dB and 9.58~dB under the D1 and D2 metrics, respectively, while providing $-93.95\%$ and $-89.42\%$ BD-BR. For attributes, DUGAE obtained BD-PSNR gains of 4.23~dB, 1.99~dB, and 2.86~dB for the Y, U, and V components, with corresponding BD-BR reductions of $-66.61\%$, $-60.03\%$, and $-66.20\%$, respectively. In addition, DUGAE brought an average BD-PCQM improvement of $12.53\times10^{-3}$ with $-79.75\%$ BD-BR in terms of 1-PCQM, confirming that the proposed framework can effectively enhance both geometry structure and perceptual attribute quality for dynamic point clouds.

In addition, Table~\ref{tab:dugae_vs_ugae} reports the objective comparison results of DUGAE against UGAE. Overall, against this stronger baseline, DUGAE delivered more moderate gains. For geometry, DUGAE achieved an average $-7.60\%$ BD-BR with a 0.34~dB BD-PSNR gain on the D1 metric. For attributes, the overall improvements were relatively limited. On the Y, U, and V components, DUGAE achieved BD-PSNR gains of 0.07~dB, 0.23~dB, and 0.25~dB, corresponding to BD-BR values of $-2.58\%$, $-16.16\%$, and $-12.78\%$, respectively. Notably, the gains on the U and V components were substantially larger than those on the Y component, likely because the Y component contains more fine-grained details and high-frequency content, making it harder to extract sufficient information from adjacent frames during spatiotemporal fusion. In terms of perceptual quality, DUGAE achieved an average BD-PCQM improvement of $0.17\times10^{-3}$ with a corresponding BD-BR reduction of $-3.43\%$ on $1-\mathrm{PCQM}$, further confirming that DUGAE can effectively exploit temporal information in dynamic point clouds.

\begin{figure*}[t]
\centering
\includegraphics[width=\linewidth]{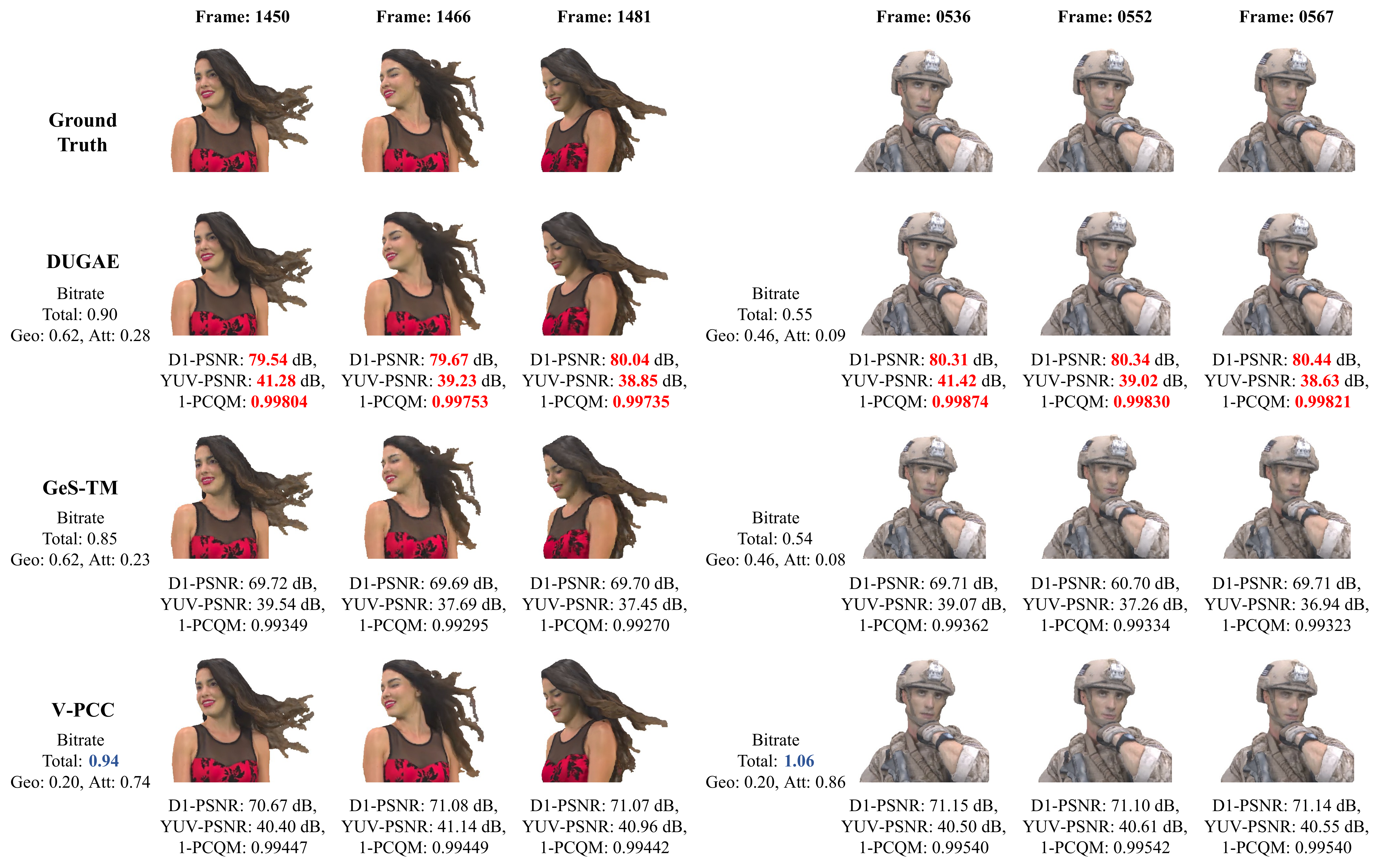}
\caption{Comparison of visual reconstruction quality. The bitrates are expressed in bpip.}
\label{fig:Subjective}
\end{figure*}

We further compared DUGAE with V-PCC, where H.265/HEVC in a low-delay configuration used for video coding. Owing to the substantial differences between the coding frameworks of V-PCC and GeS-TM, we separately evaluated geometry and attribute performance separately. Fig.~\ref{fig:R-D_G} compares the performance of DUGAE, V-PCC, GeS-TM, UGAE, and DPCGC in terms of D1-PSNR versus geometry bitrate. GeS-TM yielded the worst geometry quality, with the lowest D1-PSNR at all bitrate points. DPCGC significantly improved over GeS-TM but was still clearly inferior to the enhancement-based approaches on most sequences. In contrast, the curves of UGAE and DUGAE consistently lied at the top. DUGAE outperformed UGAE on all sequences, showing that temporal geometry enhancement can effectively exploit inter-frame correlations. V-PCC provided better geometry quality than GeS-TM but was outperformed by DUGAE and UGAE for most bitrates, particularly at low and medium bitrates where the benefit of the proposed enhancement was more pronounced. Across all sequences, DUGAE provided stable gains in D1-PSNR over the anchor codecs.

Fig.~\ref{fig:R-D_A} compares the rate-distortion performance of DUGAE, V-PCC, GeS-TM, and UGAE in terms of Y-PSNR versus attribute bitrate for the same dynamic sequences. For all sequences, GeS-TM generally provided the lowest Y-PSNR. In contrast, both UGAE and DUGAE improved attribute reconstruction quality, achieving higher Y-PSNR over the tested bitrate range. The R-D curves of DUGAE and UGAE were closely matched, with DUGAE consistently delivering slightly higher Y-PSNR, indicating that exploiting inter-frame information benefits attribute enhancement for dynamic point clouds. Compared with V-PCC, DUGAE achieved comparable, and in several cases higher, Y-PSNR within the overlapping bitrate region, particularly on \textit{Redandblack}, \textit{Ricardo}, and \textit{Phil}. In the supplemental materials, we show the variation of D1-PSNR and Y-PSNR as a fuinction of frame number. 

\begin{table}[t]
  \centering
  \caption{AVERAGE PROCESSING TIME PER FRAME for Different Components of DUGAE.}
  \label{tab:dugae_time}
  \begin{tabular}{lccc}
    \toprule
    Point Cloud & DGE (GPU) (s) & DGE (CPU) (s) & DAE (GPU) (s) \\
    \midrule
    \textit{Redandblack} & 4.90 & 30.46  & 1.69 \\
    \textit{Soldier}     & 5.29 & 42.03  & 1.76 \\
    \textit{Exercise}    & 4.97 & 122.74 & 1.67 \\
    \textit{Model}       & 5.51 & 117.41 & 1.81 \\
    \textit{Phil}        & 5.50 & 63.83  & 1.71 \\
    \textit{Ricardo}     & 6.05 & 36.92  & 1.81 \\
    \textit{Sarah}       & 4.49 & 54.77  & 1.78 \\
    \midrule
    \textbf{Average}     & \textbf{5.24} & \textbf{66.88}  & \textbf{1.75} \\
    \bottomrule
  \end{tabular}
\end{table}

\subsection{Visual Quality Evaluation}
For visual comparison, we selected three frames from two sequences (Fig. \ref{fig:Subjective}). DUGAE and GeS-TM are compared at about the same geometry and attribute bitrates, whereas V-PCC used a lower geometry bitrate and a higher attribute bitrate. In terms of geometry, DUGAE was more accurate than both GeS-TM and V-PCC. For the Redandblack sequence, the contours enhanced by DUGAE were clearer, while GeS-TM exhibited noticeable geometry roughness and V-PCC introduced redundant and missing regions (e.g., hair). Similarly, for the Soldier sequence, the helmet edges and the fine structures around the straps reconstructed by DUGAE were closer to the original point cloud. In terms of attributes, DUGAE also produced cleaner and sharper colors than GeS-TM, with few visible artifacts in highly textured regions such as clothing and facial areas. Compared with V-PCC, DUGAE avoided the common texture misalignment in projection-based reconstructions (e.g., ears, neck).

\begin{figure}[t]
\centering
\includegraphics[width=\linewidth]{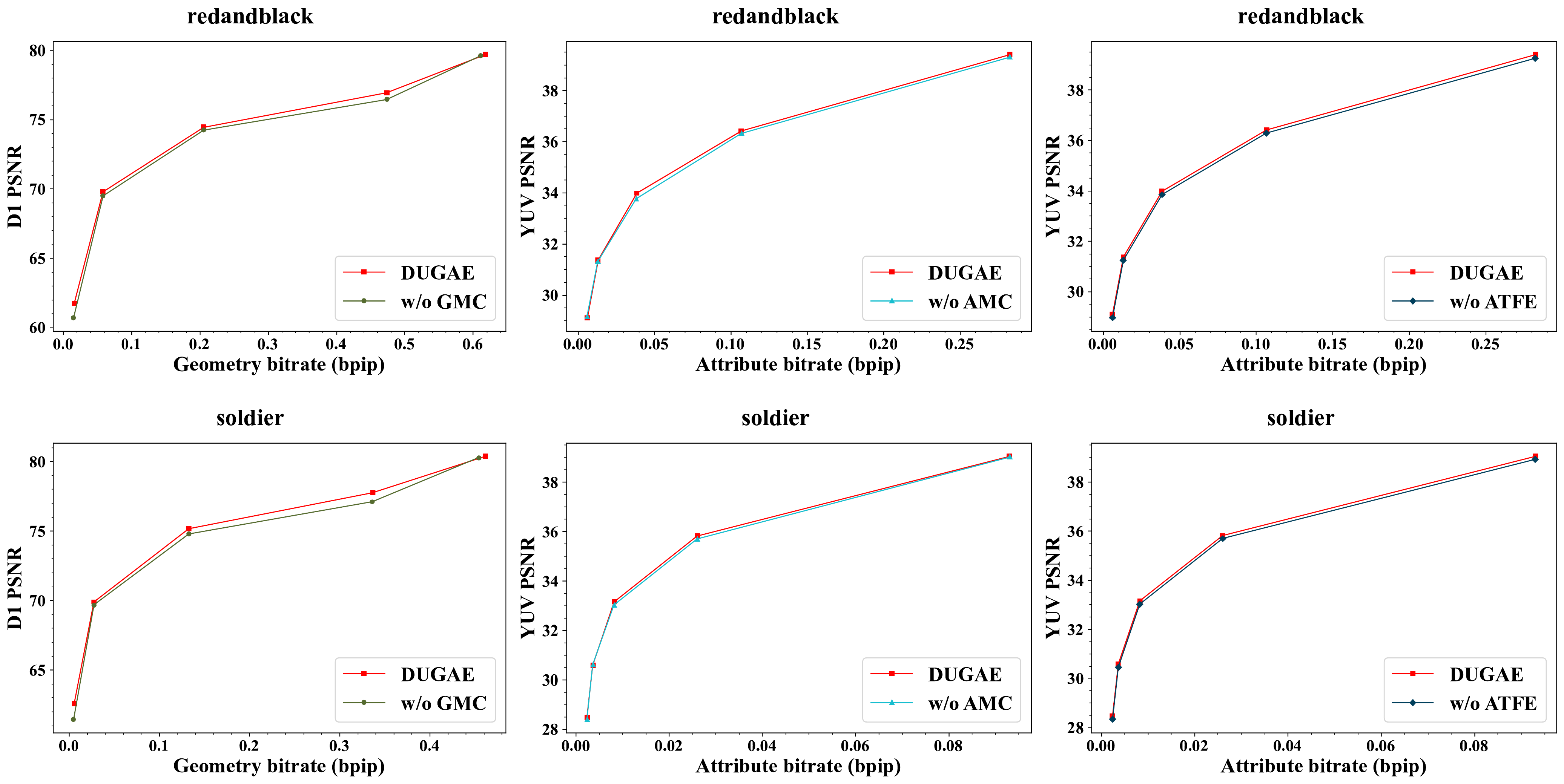}
\caption{R-D Curves for DUGAE with and without the GMC, AMC, and ATFE components on Redandblack and Soldier Dynamic Point Cloud Sequences.}
\label{fig:ablation}
\end{figure}

\begin{figure}[t]
\centering
\includegraphics[width=\linewidth]{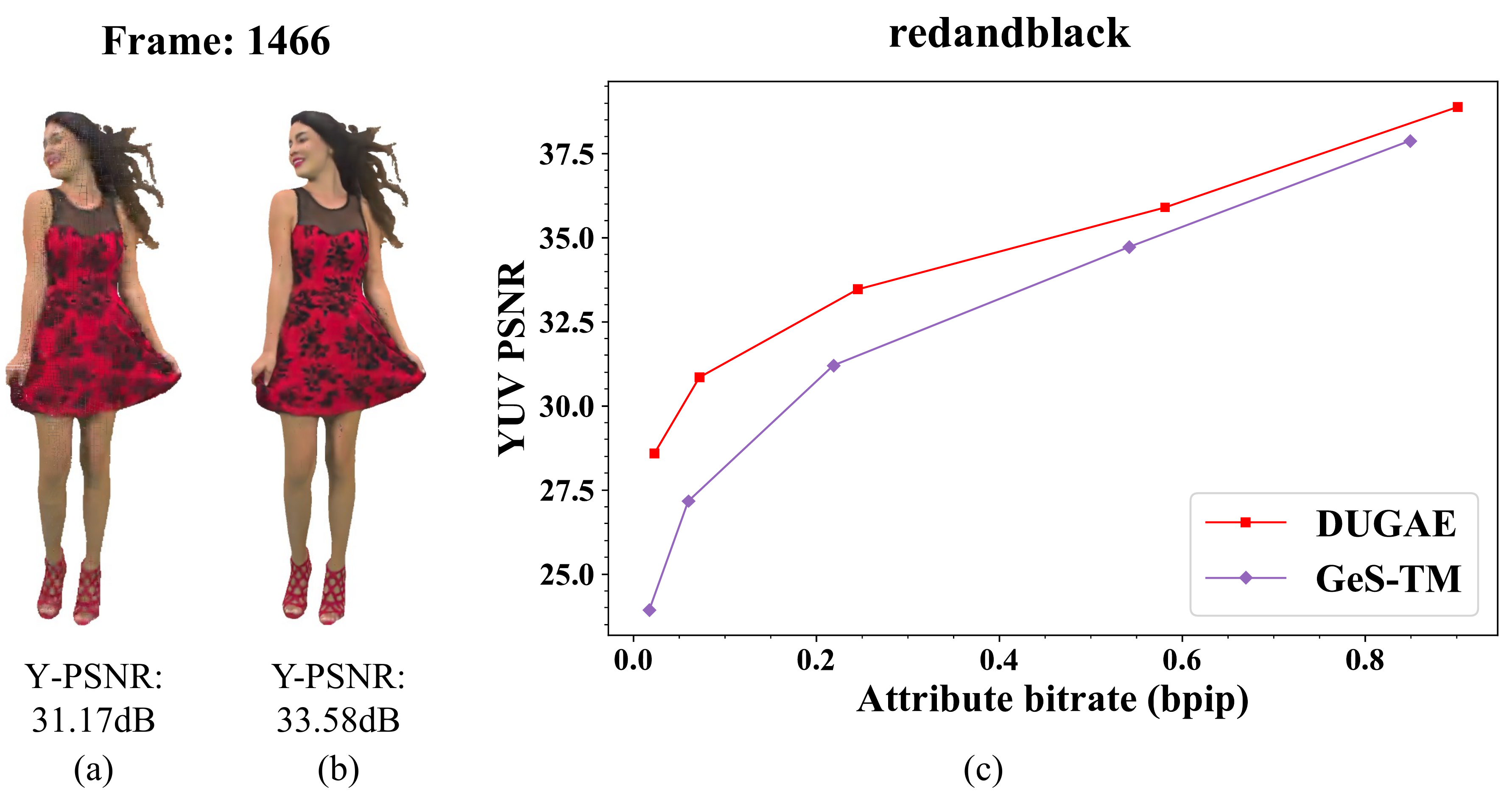}
\caption{Subjective quality comparison and R–D curves for Redandblack. From left to right: (a) GeS-TM reconstructed attributes using compressed geometry (GeS-TM), (b) GeS-TM reconstructed attributes using enhanced geometry (DUGAE), (c) R-D curves of DUGAE and GeS-TM. }
\label{fig:redandblack}
\end{figure}

\subsection{Time Complexity}
Table \ref{tab:dugae_time} reports the average per-frame processing time. For geometry enhancement, DGE-Net ran in 5.24 s/frame on GPU, compared with 66.88 s/frame on CPU. DAE-Net was substantially faster on GPU, requiring 1.75 s/frame on average. Although the GPU implementation of DGE-Net offers a large speedup, bitwise-repeatable outputs are not guaranteed because some GPU operations used in the pipeline can be non-deterministic (e.g., due to parallel reduction order and low-level kernel behavior). To ensure run-to-run consistent geometry enhancement, we used the CPU-based DGE-Net for all reported geometry results and provide the GPU runtime only for reference. In contrast, we do not enforce strict determinism for attribute enhancement, so DAE-Net was evaluated with its GPU implementation. We prioritized determinism for geometry because run-to-run variations can alter point correspondences and downstream metric computations. For attribute enhancement, minor numerical differences have negligible impact on reported metrics. 

\subsection{Ablation Study}
Fig.~\ref{fig:ablation} presents an ablation study of GMC, AMC, and ATFE on the \textit{Redandblack} and \textit{Soldier} sequences. We evaluated three variants: ``w/o GMC'', where GMC was removed and geometry was enhancedframe-by-frame; ``w/o AMC'', where AMC was removed and attributes were enhanced using only single-frame information; and ``w/o ATFE'', where ATFE was removed and both temporal and spatial attribute features were extracted by the Attribute Feature Extractor (AFE).

For the D1-PSNR metric, DUGAE achieved slightly higher geometry quality than ``w/o GMC'' at all bitrate points, showing the benefits of exploiting inter-frame correlation. For YUV-PSNR metric, DUGAE also outperformed ``w/o AMC'', confirming thebenefit of AMC for exploiting temporal redundancy in attribute enhancement. In the third column, ``w/o ATFE'' shows a clear drop in YUV-PSNR compared with the full model. This suggests that relying on a single AFE to extract both temporal and spatial features weakens the network’s capacity to learn complementary cues across frames, which leads to degraded attribute reconstruction

Furthermore, we analyzed the effectiveness of the enhanced geometry for attribute detail preservation. After recoloring the original attributes $\mathbf{A}_t$ onto the enhanced geometry $\hat{\mathbf{G}}_t$ using DA-KNN, we used GeS-TM (RAHT) for lossy coding of the recolored attributes $\tilde{\mathbf{A}}_t$ together with $\tilde{\mathbf{A}}_{t-1}$ from the previous compressed frame. As shown in Fig.~\ref{fig:redandblack}, reconstructions from GeS-TM tended to lose high-frequency attribute detail, resulting in blurred textures and coarse boundaries. In contrast, DUGAE, benefiting from the enhanced geometry and DA-KNN, yielded more distinguishable color details and visually sharper boundaries. The R-D curves in Fig. \ref{fig:redandblack} (c) further show that DUGAE provided a substantial gain with only a small bitrate increase, with the improvement most pronounced in the low-bitrate region.

\section{Conclusion}
\label{sec:conclusion}
We proposed DUGAE, a unified framework for geometry and attribute enhancement of G-PCC compressed dynamic point clouds. DUGAE decomposes enhancement into three sub-tasks: dynamic geometry enhancement, attribute recoloring, and dynamic attribute enhancement, implemented by DGE-Net, DA-KNN, and DAE-Net, respectively. DGE-Net uses an SPConv-based U-Net with GMC to align temporal features. DA-KNN recolors the original attributes onto the enhanced geometry at the encoder side, improving mapping completeness and preserving details. DAE-Net further refines attributes with ATFE and AMC by exploiting spatial and temporal information. Experiments on dynamic point cloud sequences show that DUGAE consistently outperforms state-of-the-art methods in both geometry and attributes. Future work may consider designing lightweight architectures, and adapting the proposed framework to a wider range of point cloud codecs.

\end{document}